\documentclass[conference]{IEEEtran}
\IEEEoverridecommandlockouts
\usepackage{cite}
\usepackage{amsmath,amssymb,amsfonts}
\usepackage{algorithmic}
\usepackage{graphicx}
\usepackage{textcomp}
\usepackage{xcolor}
\def\BibTeX{{\rm B\kern-.05em{\sc i\kern-.025em b}\kern-.08em
    T\kern-.1667em\lower.7ex\hbox{E}\kern-.125emX}}

\makeatletter

\def\ps@IEEEtitlepagestyle{%
	\def\@oddfoot{\mycopyrightnotice}%
	\def\@evenfoot{}%
}
\def\mycopyrightnotice{%
	{\footnotesize  979-8-3315-3254-3/24/\$31.00 \textcopyright 2024 IEEE\hfill}
	\gdef\mycopyrightnotice{}
}
\usepackage{amsmath}
\usepackage{amssymb}
\usepackage{arydshln}
\usepackage{subcaption}
\usepackage{mathtools}
\usepackage{nccmath}
\usepackage{adjustbox}
\usepackage{multirow}
\usepackage{tikz}

\usepackage{afterpage}

\usepackage[compact]{titlesec} 
\titlespacing{\section}{0pt}{*0.9}{*0.5}

\newcommand\VspaceImages{0}

\usepackage{cite}
\usepackage{amsmath,amssymb,amsfonts}
\usepackage{algorithmic}
\usepackage{graphicx}
\usepackage{textcomp}
\usepackage{xcolor}
\usepackage{comment}
\def\BibTeX{{\rm B\kern-.05em{\sc i\kern-.025em b}\kern-.08em
    T\kern-.1667em\lower.7ex\hbox{E}\kern-.125emX}}

\makeatletter

\def\ps@IEEEtitlepagestyle{%
	\def\@oddfoot{\mycopyrightnotice}%
	\def\@evenfoot{}%
}
\def\mycopyrightnotice{%
	{\footnotesize  979-8-3315-3254-3/24/\$31.00 \textcopyright 2024 IEEE\hfill}
	\gdef\mycopyrightnotice{}
}



\makeatletter
\newcommand*\titleheader[1]{\gdef\@titleheader{#1}}
\AtBeginDocument{%
	\let\st@red@title\@title%
	\def\@title{%
		\bgroup\normalfont\large\centering\@titleheader\par\egroup
		\vskip0.5em\st@red@title}
}
\makeatother
\title{Viability of Robot-supported Flipped Classes in English for Medical Use Reading Comprehension
}
\usepackage{array}
\usepackage{makecell}

\titleheader{
	\vspace*{-1.9cm}
	\begin{center}
			\begin{tabular}{m{0.09\textwidth} m{0.7\textwidth} m{0.09\textwidth}} 
				\centering\includegraphics[width=0.07\textwidth]{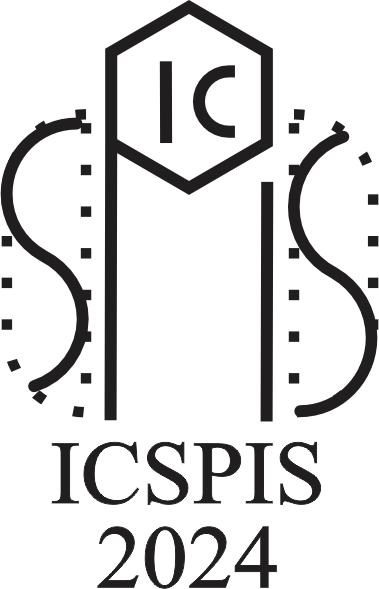} &
				\centering\raisebox{0.5em}{
					\makecell[c]{
						\normalsize
						10\textsuperscript{th} International Conference on Signal Processing \& Intelligent Systems \\ \normalsize
						Shahrood University of Technology, December 25-26, 2024
					}
				} &
				\centering\includegraphics[width=0.07\textwidth]{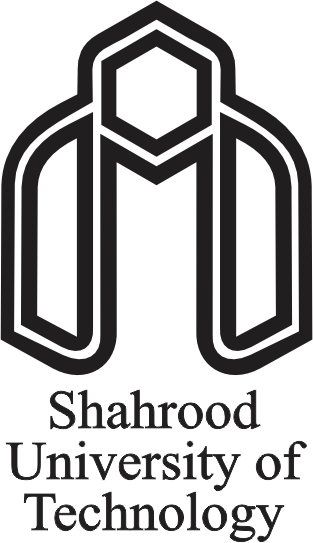}
			\end{tabular}
		\end{center}
	}

\begin{document}

\author{\IEEEauthorblockN{Amin Rezasoltani$^\star$}
\IEEEauthorblockA{\textit{Ava and Nima Social Robotics} \\
\textit{Dr. Robot Company}\\
Tehran, Iran \\ 
aminrezasoltani123@gmail.com}
\and
\IEEEauthorblockN{Ehsan Saffari}
\IEEEauthorblockA{\textit{Ava and Nima Social Robotics} \\
\textit{Dr. Robot Company}\\
Tehran, Iran \\ 
e$\_$saffari@yahoo.com}
\and
\IEEEauthorblockN{Farzam Tajdari}
\IEEEauthorblockA{\textit{Mechanical Engineering} \\
\textit{Delft University of Technology}\\
Delft, Netherlands \\
f.tajdari@tudelft.nl}
}

\maketitle

\begin{abstract}

This study delved into the viability of Robot-supported flipped classes in English for Medical Purposes reading comprehension. In a 16-session course, the reading comprehension and then workspace performance of 444 students, with Commercially-Off-The-Shelf and Self-Generated robot flipped classes were compared. The results indicated that the flipped classes brought about a good instructional-learning ambience in postsecondary education for English for Medical Purposes (EMP) reading comprehension and adopting proactive approach for workspace performance. In tandem, the Mixed Effect Model revealed that student participation in the self-generated robot-supported flipped classes yielded a larger effect size (+17.6\%) than Commercially-Off-The-Shelf robot-supported flipped classes. Analyses produced five contributing moderators of EMP reading comprehension and workspace performance: reading proficiency, attitude, manner of practicing, as well as student and teacher role.

\end{abstract}


\section{Introduction}
\label{sec:Introduction}

Researchers \cite{bremner2018bringing}, \cite{lo2011esp} state that heavy dependency on less flexible conventional approaches for teaching English for Specific Purposes (ESP) is the major cause of failure in teaching ESP reading comprehension. Sherwood in \cite{sherwood1994first} contends, "what is now required [in ESP instructional-learning contexts] is a consideration of the call for more transferable skills in the light of contemporary figures on the employment of graduates". These results lead to the importance of providing hands-on lessons to actively involve the students in the teaching process.

ESP courses follow two major aims of preparing collegiate students for their academic life (viz., English for Specific and Academic Purposes or ESAP) and future career in post-academic contexts (viz., English for Specific and Occupational Pur-poses or ESOP), hence, current and future working conditions demand that teaching ESP reading should not be restricted only to the classroom \cite{javid2015english}. Students are in need to not only understand the academic materials, but also to communicate with their cohorts in the international working milieus \cite{chen2019using}.

Along these lines, educational technology (EdTech) devotees are looking to what is to come \cite{noonoo2014google}. EdTech has changed the way individuals access information, and communicate. EdTech-supported language education has brought about rich opportunities for communication and efforts among students; this way, EdTech-supported English language education has been successful in increasing the motivation of students to develop their understanding. It is increasingly clear that EdTech represents Language for Specific Purposes (LSP) more than a helpful realia.

Robots~\cite{tajdari2023adaptive, tajdari2021discrete, tajdari2017design, tajdari2017robust, tajdari2017switching, tajdari2020intelligent, tajdari2022implementation} functioning as teaching assistants present a chance to enrich education \cite{li2020study, chang2024robot}. Researchers have suggested that robots can act as teaching assistants, learning companions, or learning tools to enhance the academic setting \cite{johnson2018cooperative, kennedy2016social}. A learning framework supported by robot teaching assistants provides tailored assistance during instructor-guided lessons, aiding in the understanding of difficult subject matter \cite{engwall2022interaction}. Furthermore, this approach not only boosts students' academic outcomes but also positively impacts their attitudes and engagement in learning \cite{kubilinskiene2017applying}.

Despite the influence that robots have continued to exert on the LSP, most studies in RBLE have only been carried out in a small number of areas. This lack is even more profound when it comes to postsecondary education. This study intended to examine the viability of robot-supported flipped classes in teaching ESP reading comprehension (with special reference to Medical English, namely EMP), and to identify the factors that are important to employing robot-supported flipped classes of postsecondary education.

\section{Method}
\label{sec:Methods}

The study was done in two phases: in the first phase we were focused on designing and manufacturing a proper robot for the current study, and in the second phase we arrange a questionnaire survey on students in robot-supported flipped classes.

\subsection{"Safir" Robot}
\label{subsec:Safir}

\begin{figure}
\centering
\includegraphics[width=0.6\linewidth]{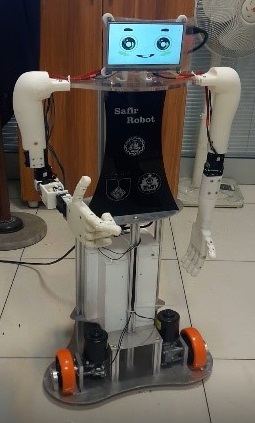}
\caption{The Safir Robot.} \label{fig:safir}
\vspace{\VspaceImages cm}
\end{figure}

A new robot was designed and manufactured for the planned robot-supported classes as shown in Fig.~\ref{fig:safir}. The robot was a 1 meter height humanoid mobile robot. It has 10 Degrees of Freedom (DoFs): 2 in base, one in neck, one in left hand and 6 in the right hand. These DoFs let robot to move in the class on its wheels; nodding using its head; raising both of its 3D printed hands~\cite{tajdari2023advancing, tajdari20244d, tajdari2023non, tajdari2022optimal, tajdari2022dynamic, tajdari2022next, yang2021posture, minnoye2022personalized, tajdari2022feature, tajdari2021imageprediction}; pointing and making fist using its right hand with its 2 DoFs cable driven fingers. The robot also has an LCD Face and a speaker which let it to read aloud preprogrammed texts and show the proper lip motions.

The main processor of Safir is a Raspberry Pi board. A python program was written for Safir to allow the user to control it using a remote keyboard~\cite{tajdari2023optimal, tajdari2022flow, tajdari2023online, tajdari2020intelligentcontrol, tajdari2021simultaneouschanger, tajdari2019fuzzy, tajdari2019integrated, tajdari2021adaptive, tajdari2020feedback, tajdari2024perception, tajdari2024NonRigid}. It could play a recorded audio of a reading, moving the robot's lips and moving robots hands, fingers and head based on a YAML file. This way we could store multiple scenarios for the classes.

\subsection{The Survey}
\label{subsec:The Survey}

In the academic year 2017-2018, a sample of 444 students (140 males and 304 fe-males from the disciplines of Medical Library, Nursing, Health Information Technology (HIT), Nutritional Sciences, and Pharmacy) from as many as 463 students of Medical University of Isfahan was selected. They were in EMP courses. Further, four TEFL major students who took compulsory course of Materials Development were randomly selected to be included as members of the self-generated circles. 

To identify their levels of proficiency, they took 30 items of Test of English as a Foreign Language (TOEFL)-like reading (5 passages each with 6 multiple-choice items). The test enjoyed the reliability of r=.79 as well as content and face validity.

\begin{figure}
\centering
\includegraphics[width=1\linewidth, height = 0.6\linewidth]{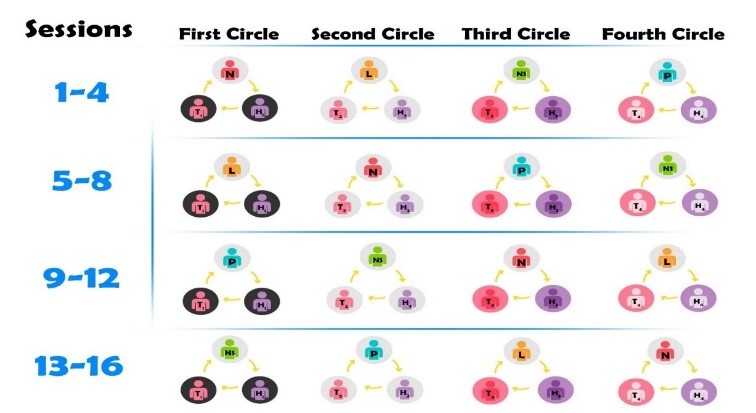}
\caption{The make-up of circles.} \label{fig:circles}
\vspace{\VspaceImages cm}
\end{figure}

To practice the materials under the surveillance of subject-area and English teachers, the students were randomly assigned to two sets of Commercially-Of-The-Shelf- (n=340) and self-generated- (n=104) AR-supported activities. Students in the Commercially-Of-The-Shelf set were further randomly assigned to either individual (n=220) or collective group (n=120). As for the collective practicing, students were divided into three-member circles. Students in the self-generated set were randomly divided into the monodisciplinary (n=96) and interdisciplinary groups. A shot of the interdisciplinary circles is shown in Fig.~\ref{fig:circles}.

As to developing self-generated activities in monodisciplinary group circles, the students were invited to complete 64 templates. The details about the participants are summarized in Table~\ref{tab:table1}. In the table, please note that Subject is equal to subject-area teacher; to explore the degree of success that might be achieved by the teachers, while half of the participants were randomly assigned to classes that were conducted by an English teacher, the other half assigned to classes that were conducted by a subject-area teacher.
\begin{table}[tb]
	\caption{the correlation between the hidden variables (Fornel and Locker Analysis)}
	\label{tab:table1}
	\centering
	\includegraphics[width=\linewidth]{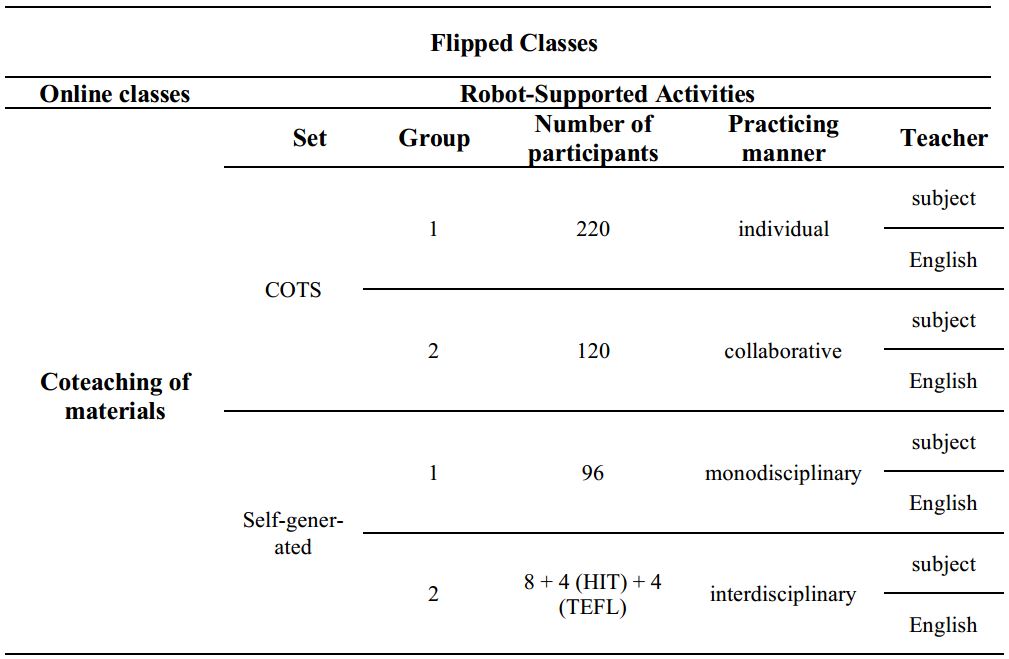}
 \vspace{\VspaceImages cm}
\end{table}

This complementarity study with full factorial design was conducted as follows:
\begin{itemize}
    \item \textbf{Step I:} Initially, the students were given the Google form of a Persian attitude questionnaire with 21 items covering the five categories of EMP teaching through robot-supported classes. To identify students' Basic Technology (BTC) levels, the second part of the questionnaire was assigned to students' self-assessment. The questionnaire was face and content validated by five TEFL and subject-area experts. It was tested for the reliability using Cronbach's Alpha (r=0.8). A chi-square over the degree of freedom value, $X^2/df$=2.34 showed a good fit. 
    
    \item \textbf{Step II:} Treatment and assessment: This study was conducted in 16 sessions. To conduct the flipped classroom in addition to the 90-minute-a-week online sessions, 90-minute-a-week sessions for practicing robot-supported activities were added. 
\end{itemize}

The materials were taken from \textit{English for the Students of Nutrition} \cite{akbari2020iran}, \textit{English for the Students of Pharmacy} \cite{alizadeh2021investigation}, \textit{English for the Students of Nursing} \cite{teo2022exploring}, \textit{English Texts for the Students of Library} \cite{anggraini2022interaction}. The materials were changed into computer-readable passages for online sessions. 

Regarding the self-generated activities, a mini-corpus containing 240 subject-area passages was developed. Students selected materials from the mini-corpus. Consequently, the prototypes of the activities were developed to be practiced by the students in the subsequent sessions. Fifty-six Commercially-Of-The-Shelves were selected for Safir robot. Their motifs were in line with the topic of the lessons.

\begin{itemize}
    \item \textbf{Step III:} Assessment of students' performance in real-world workspace: Six weeks after the final session, students arrived to the workspaces to be assessed regarding their ability to use the materials.

    \item \textbf{Step IV:} Interview: As the assessment of students' performance in real-world work-spaces was completed, interview with the students with the lowest and highest scores from each group was conducted in Persian. The interviews were transcribed and analyzed by the researchers. 
\end{itemize}

\section{Results}
\label{sec:results}

\subsection{Students' responses to the questionnaire}

Analyzing the students' responses revealed that most of them were dissatisfied with the features of common English classes. They complained that in Higher Education Institutions (HEIs), students learn how to practice textbook materials and exercises but did not get any coursework in how to tap into the materials effectively in the actual world. Martín et al. \cite{del2015textbooks} state that these exercises underspecify the real-world features of materials. A great majority of the students held positive or fairly positive attitudes towards the extended learning of reading English. Playful practicing of reading could help students reevaluate the information in the materials \cite{freeman2017case}. Accordingly, a substantial minority voted for reliance only on textbook exercises. 

When they were asked if they prefer activities to be related to their daily lives, they all with one accord said that what they liked best was doing well as workforce. When it comes to integrating robot across subjects to boost student English learning, a majority of the students were of the opinion that simulation of real-world through robots can foster wider outreach among students. Additionally, they opined that reading classes need to be presented in authentic contexts. A similar population believed that these activities better cater to students' needs. They highlighted the need for functioning adequately in both academia and workspaces. They said that robots help to foster more accurate mental representations of the materials. Through simulating the workspaces, students can be well disposed to give their attention to main points in the subject-matter area \cite{anthony2018introducing}. However, meanwhile, they vetoed the idea that robot-supported activities can facilitate learning of different EMP skills to the same extent. The results were in line with the Smith's assumption in \cite{smith2000attitudes} that different platforms "impart different skills to students and certain skills are more attractive to some employers than others" (p. 282).

Most of the students asserted that students need to be valued members of the in-structional-learning setting. They preferred to become involved rather than observe the workspaces.  Boyne et. al \cite{boyne1992enterprise} suggest ways of es-tablishing engaging contexts by fostering "a range of more active, experiential, stu-dent-centered approaches to learning, especially in conjunction with the workplace, would be likely to produce the desired enterprise outcomes" (p. 6). This active role gives students suffrage to choose activities consistent with their needs. From the respondents' per-spectives, practicing robot-supported reading in interdisciplinary milieu presages qualifications of comprehension. Under such circumstances, students can collaborate with each other as well, thus, they will come up with answers on their own. The pro-posal for coteaching in robot-supported teaching milieus was held in tight embrace of the respondents. According to \cite{durlak2011impact}, as interest in engaging students continues to grow, ease of use, as the prominent feature of robots, can play a big part in helping students become more involved in their academic and future lives. 

\subsection{Analyzing the students' reading comprehension and performance}

For data analysis, Linear Mixed-Effect Model with random intercept and random slope was used. As shown in Table~\ref{tab:table2}, teaching materials through Commercially-Of-The-Shelf robot-supported flipped classes did little to encourage students to stand on their own feet (MCOTS.IN.E = 14.6 \& MCOTS.COLL.E = 16.41). The rate of progress was the highest when the self-generated activities were practiced in interdisciplinary (M16E = 19.75) vs. monodisciplinary circles (M16E =17.58). The frequency of incomplete activities in online classrooms and lack of success in the real-world scenes could be indication of the lack of students' ability who practiced through Commercially-Of-The-Shelf vs. self-generated activities (MRE.COTS = 14.6, MRE.SG. = 19). Of course, the rate of progress was different in the groups as far as the teacher role was concerned to the extent that the students who were taught by English teachers achieved greater progress (ME.COTS = 16.41$>$MSUB.COTS = 16.31; ME.SG = 19.75$>$MSUB.SG = 17.75) and workspace score than their counter-parts who were taught by the subject-area teacher (MEREAL = 14.43$>$MSUBJREAL = 14.2; MEREAL = 19$>$MSUBJREAL= 17.75).
\begin{table}[tb]
	\caption{Comparison of the Participants' Progress.}
	\label{tab:table2}
	\centering
	\includegraphics[width=\linewidth]{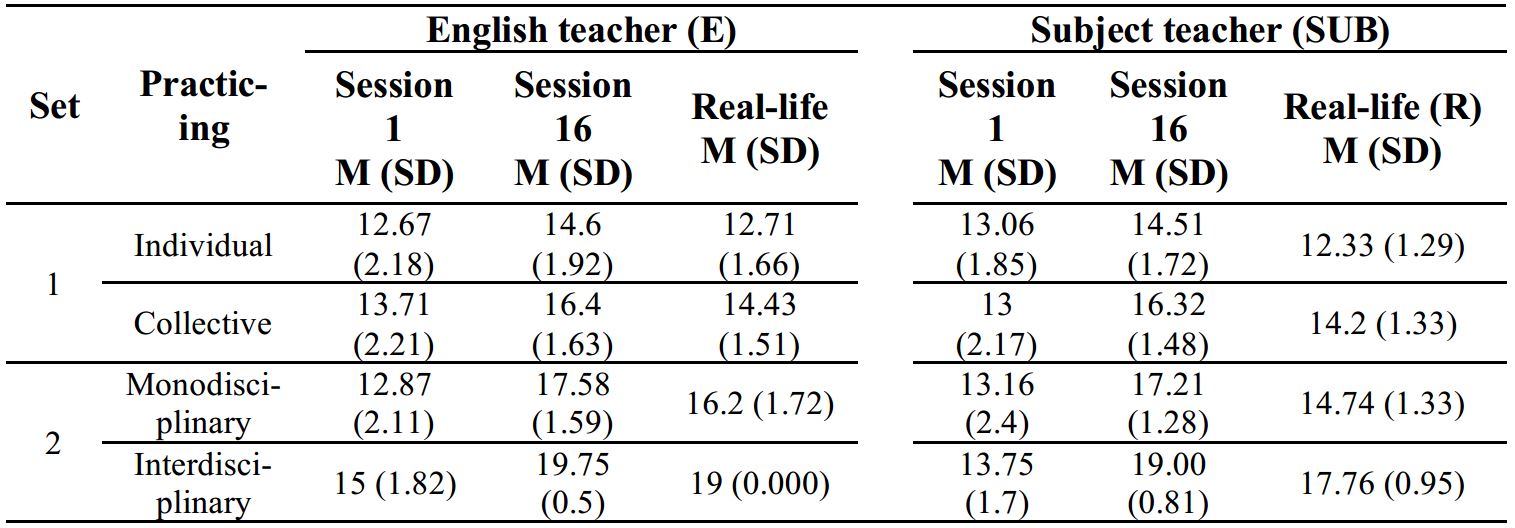}
 \vspace{\VspaceImages cm}
\end{table}

As to inferential analysis of the data, baseline scores were considered as covari-ate; thus, the participants' scores were adjusted regarding their different baseline scores.

Table of Hypothesis Testing for Within-Subject and Between-Subject effects on students' reading comprehension (Table~\ref{tab:table3}) shows the probable effects of within-subject and between-subject parameters on students' comprehension and performance.
\begin{table}[tb]
	\caption{Hypothesis Testing for Within-subject and Between-subject Effects on the Students' Com-prehension and Performance.}
	\label{tab:table3}
	\centering
	\includegraphics[width=\linewidth]{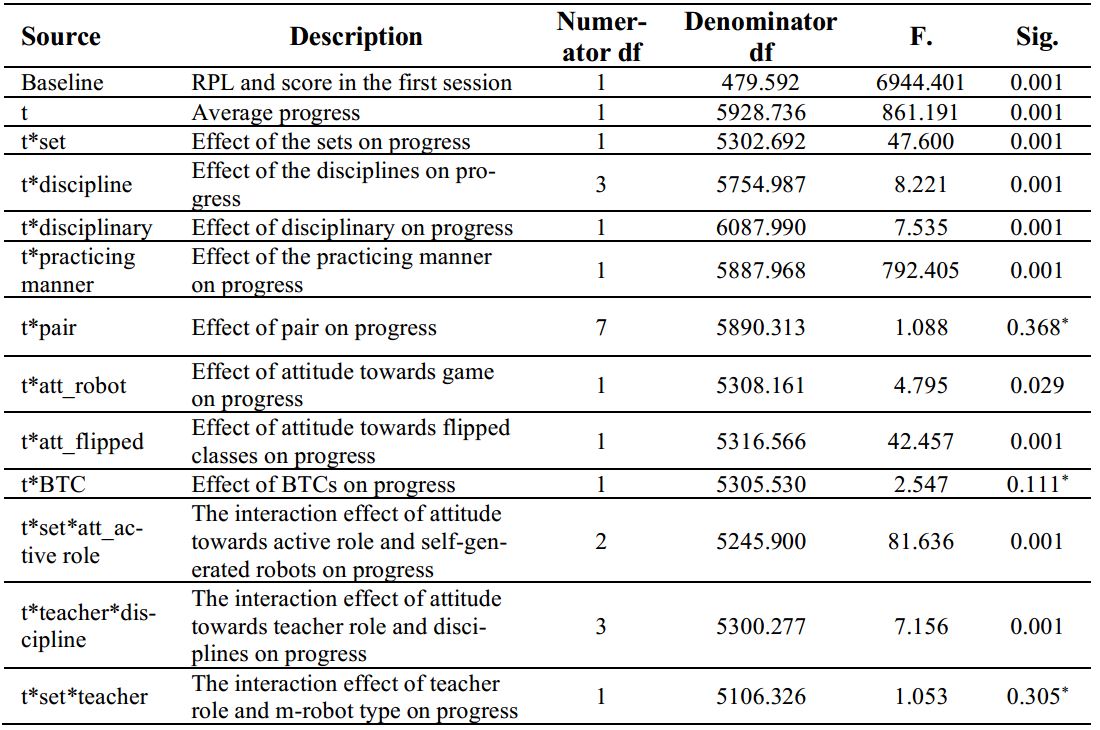}
 \vspace{\VspaceImages cm}
\end{table}

Analysis of the data shows the significant effect of reading proficiency (RP) on students' EMP reading in the first session (F = 6944.401, Sig. = 0.001). The effect of using flipped classes for teaching EMP reading comprehension on progress rate was significant (F = 861.191, Sig. = 0.001). t$^{\star}$set shows that students' progress was significantly different in the sets (F = 47.600, Sig. = 0.001). t$^{\star}$discipline shows that students' progress was significantly different in different disciplines (F = 8.221, Sig.= 0.001). Similarly, t$^{\star}$disciplinary discloses that students' progress was significantly different in monodisciplinary and interdisciplinary groups (F = 7.535, Sig. = 0.006). But, t$^{\star}$pair shows that students' movement in interdisciplinary circles with different HIT and TEFL members did not give rise to a significant differences in their EMP comprehension (F = 1.088, Sig. = 0.368). t$^{\star}$att$\_$robot indicates that students' attitudes towards robot-supported activities did not lead to significant difference in their progress (F = 4.795, Sig. = 0.029). Also, students' BTC did not bring about significant differences in their progress (F = 2.547, Sig. = 0.111). t$^{\star}$att.$\_$flipped classes, however, indicates that the students' attitudes towards robot produced significant differences in their EMP reading (F = 42.457, Sig. = 0.001). t$^{\star}$set$^{\star}$att$\_$active role shows there is an interaction effect between practicing self-generated activities and attitude towards playing active role in these modules on progress of students (F = 81.363, Sig. = 0.001). t$^{\star}$teacher$^{\star}$discipline reveals the interaction effect between teacher and students' discipline on their EMP comprehension (F = 7.156, Sig. = 0.000); however, this was not the case as far as the interaction effect of teacher and robot types on students' comprehension was concerned (F = 1.053, Sig. = 0.305).

To compare the participants' scores in workspaces, the emphasis was put on comparing longevity in learning. For that reason, the students' workspace scores were adjusted lest the effect of final session is present; put simply, the effect of teaching and longevity was separated. This way, the students' scores in the last session were co-variated and their workspace scores were adjusted; thus, the adjusted scores, void of teaching, practicing, and learning effect, were taken into account. According to the table of the Analysis of variances (Table~\ref{tab:table4}) and the table of Differences in the Progress in Table~\ref{tab:table5}, the participants' workspace scores were predictable from their final session scores (viz., the effect of teaching, practicing, \& learning). Although the participants' discipline (F = 1.446, Sig. = .218) did not lead into significant differences in terms of longevity, set (F= 78.174, Sig. = 0.000), practicing manner (F = 18.232, Sig. = .000), teacher type (F = 31.492, Sig. = .000), and disciplinary circles (F = 19.412, Sig. = .000) resulted in significant differences in longevity. In tandem, Partial Eta Squared of .153 revealed that activity exerted the most profound effect on longevity.
\begin{table}[tb]
	\caption{Analysis of Variances for the Participants' Workspace Performance.}
	\label{tab:table4}
	\centering
	\includegraphics[width=\linewidth, height = 0.33\linewidth]{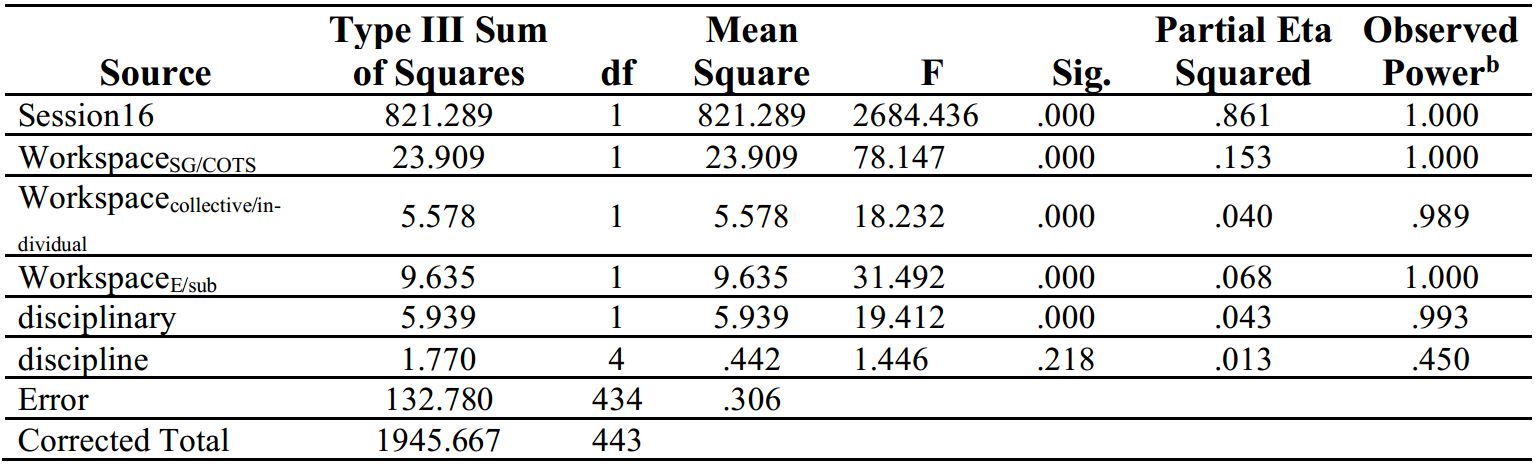}
 \vspace{\VspaceImages cm}
\end{table}
\begin{table*}[tb]
	\caption{Difference in the Progress.}
	\label{tab:table5}
	\centering
	\includegraphics[width=0.7\linewidth]{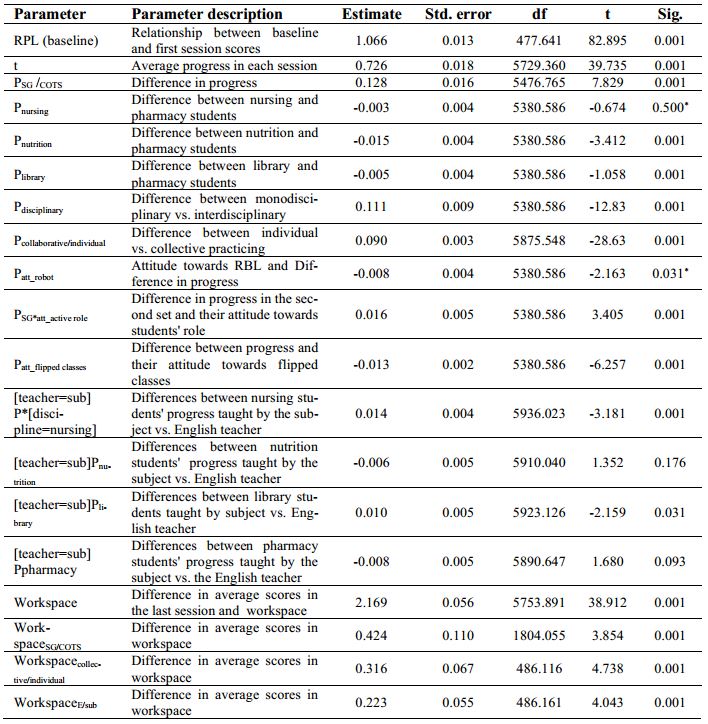}
 \vspace{\VspaceImages cm}
\end{table*}

\subsection{Students' responses to the interview}

Retrospection of the students' answers to the prompts revealed that though there were varying opinions, a set of features was evolved which were collated to the final list:
Flipped classes make it possible to enjoy the advantages of two modes of presentation and practice. The students' interpretation was that when they read EMP mate-rials from different sources their comprehension was boosted. The scenes of robot-supported activities provided repeated exposure to real language use. This helped students gain stronger educational experiences. The students' statements indicated that robot-supported activities were useful for encouraging greater participation along with practicing reading comprehension.

More interesting, perceived usefulness was more influential in the accounts of the students from self-generated groups. As said by these students, each student has specific needs and m-robot-supported activities were of help in detecting students' problems in comprehension. They knew the ability to develop personalized activities as the maximum benefit of flipped classes with self-generated activities. On their words, they could gain lots of ideas from the materials and employ them in real-world arenas. Similarly, they had ideas to make things better and self-generated project allowed them to do it. Students thought their interaction for developing activities as dialogic effort for comprehension, though they said they support each other in meeting the expectations. This way, practicing via robots helped students build future-ready learning experiences. 

Besides, students who practiced self-generated robots became more risk-taking as stated by these students. From their view, the burden of teaching shifts to certain extent to students; thus, it was a win-win situation for both teachers and students. They were of the opinion that robot-supported reading had the potential to be conducted fully online in outdoors. Nevertheless, they saw the online classroom as a precondition for the successful development of reading comprehension.

Interestingly, majority of the students, who practiced through their robots, stated that they had a good handle on components of academia and workspaces. As said by the students, when they were endowed with the right for preparing activities, scenes were set to read between the lines. Conversely, the participants from the Commercially-Of-The-Shelf group blamed the lack of taking full advantage of r-supported activities to their passive presence in the instructional-learning contexts. Even so, mention should be made that, students with low level of English reading proficiency highlighted the great novelty of self-generated activities and complained that developing activities could be considered mystifying to them. They altogether emphasized that before jumping into using robots, teachers needed adequate training. By the way, they highly endorsed the coteaching in online classrooms. Along these lines students opined that affordability and productivity of robots work round to using robots in EMP reading comprehension. The result of analyzing the responses dis-closed that many of the factors that influence the successful adoption of flipped classes are similar to the factors identified in productive academia and workspaces. The classifications constituting these elements were joined to figure a blueprint to function as a summary for flipped classes.

\section{Discussion}
\subsection{Robot-supported flipped classes and constructionism approach}
The constructivism approach describes how student active participation influences student comprehension and performance, which, in tandem, smooths the path for student proactive role. A finding disclosed from this study was that the robot-supported flipped classes in postsecondary education appeared successful for both academic and professional purposes. These classes demonstrated gains in EMP reading comprehension. Such finding was on the side of both students' attitude and perception, that is, by embracing the possibility of developing variegated activities for students that account for their needs, flipped classes held great promise for helping students in learning EMP reading materials. With respect to students' ability in tapping into their EMP reading comprehension in workspaces and their outperformance as a result of practicing EMP materials via AR-supported flipped classes it could be claimed that flipped classes have the potential to conceptualize authentic settings, namely Situated Learning Theory. Along with Brown's saying in \cite{brown1986investigating}, it was revealed that in giving students reading activities out of context we set them a difficult task. In effect, with the exquisite features of robot-supported flipped classes, students could access the materials pertaining to their academics. This, as indicated before, seems to confirm findings of studies in ESP reading comprehension to this effect that flipped classes through combining customary classrooms and EdTech furnish contexts with reinforced comprehension \cite{bremner2018bringing}, \cite{shokouhi2005new}. However, it falls in contrast to the findings of studies which have pointed to the temporary nature of learning that resulted from robot-based education \cite{hung2014effects}. The finding also echoed the implications of the previous studies that graduate students' failure in workspaces could be attributed to inefficiency of postsecondary teaching programs \cite{gray2019utilizing}. Equally, the result of this study implied that teaching related to students' actual subject-area activities, namely student engagement in the activities they are confronted with their academic and professional lives facilitate student proactive role.

\subsection{The way(s) of conducting robot-supported flipped classes}

The result of the study appeared to support the viewpoint that it would be naïve to acknowledge that the mere use of robots results in students' better comprehension and outperformance. The result was not only dependent on robots but also to innovative course of action for employing robots for teaching. The result went in for the dictum that comprehension of materials remains unchanged for students irrespective of way of practicing \cite{soltani2020augmented}. This implies that the optimal integration of ro-bot-supported activities into flipped classes occurred when these activities were students' self-generated type \cite{chen2019using}.

Again, finding reveled that self-generated AR activities could help students immerse in rich details. And, students could employ sufficient information by incorporating cues. These activities proved the efficiency of visualization as one of the reading comprehension strategies. As a matter of fact, students' active presence in the process of teaching EMP materials could ease the cognitive load. Interestingly, the difficulty level of the passages selected by the students in developing self-generated setting increased parallel with the increase in the difficulty level of the passages in online classes, that is, students' active role in developing activities heightened their awareness of the materials and contexts. This way, self-generated activities created association with other works, namely interdisciplinary learning conditions. And this is the right place that windows were opened for entrepreneurship along with education (viz. edupreneurs).

\bibliographystyle{IEEEtran}
\bibliography{IEEEabrv,mybibfile}

\end{document}